\documentclass[10pt,twocolumn,letterpaper]{article}

\usepackage{cvpr}
\usepackage{times}
\usepackage{epsfig}
\usepackage{graphicx}
\usepackage{amsmath}
\usepackage{amssymb}
\usepackage{graphicx}
\usepackage{wrapfig}

\usepackage[breaklinks=true,bookmarks=false]{hyperref}

\newcommand*{\affaddr}[1]{#1} 
\newcommand*{\affmark}[1][*]{\textsuperscript{#1}}

\cvprfinalcopy 

\def\cvprPaperID{****} 

\begin{document}

\title{FBI-Pose: Towards Bridging the Gap between 2D Images and 3D Human Poses using Forward-or-Backward Information}


\author{%
Yulong Shi\affmark[1], Xiaoguang Han\affmark[2], Nianjuan Jiang\affmark[1], Kun Zhou\affmark[1], Kui Jia\affmark[3], Jiangbo Lu\affmark[1]\\
\affaddr{\affmark[1]Shenzhen Cloudream Technology Co., Ltd.}\\
\affaddr{\affmark[2]Shenzhen Research Institute of Big Data, the Chinese University of Hong Kong(Shenzhen)}\\
\affaddr{\affmark[3]South China University of Technology}
}


\maketitle

\begin{abstract}
   Although significant advances have been made in the area of human poses estimation from images using deep Convolutional Neural Network (ConvNet), it remains a big challenge to perform 3D pose inference in-the-wild. This is due to the difficulty to obtain 3D pose groundtruth for outdoor environments. In this paper, we propose a novel framework to tackle this problem by exploiting the information of each bone indicating if it is forward or backward with respect to the view of the camera(we term it Forward-or-Backward Information abbreviated as FBI). Our method firstly trains a ConvNet with two branches which maps an image of a human to both the 2D joint locations and the FBI of bones. These information is further fed into a deep regression network to predict the 3D positions of joints. To support the training, we also develop an annotation user interface and labeled such FBI for around 12K in-the-wild images which are randomly selected from MPII (a public dataset of 2D pose annotation). Our experimental results on the standard benchmarks demonstrate that our approach outperforms state-of-the-art methods both qualitatively and quantitatively.
\end{abstract}

\section{Introduction}
\label{sec:intro}
\begin{figure}
  \centering
  \includegraphics[width=0.45\textwidth]{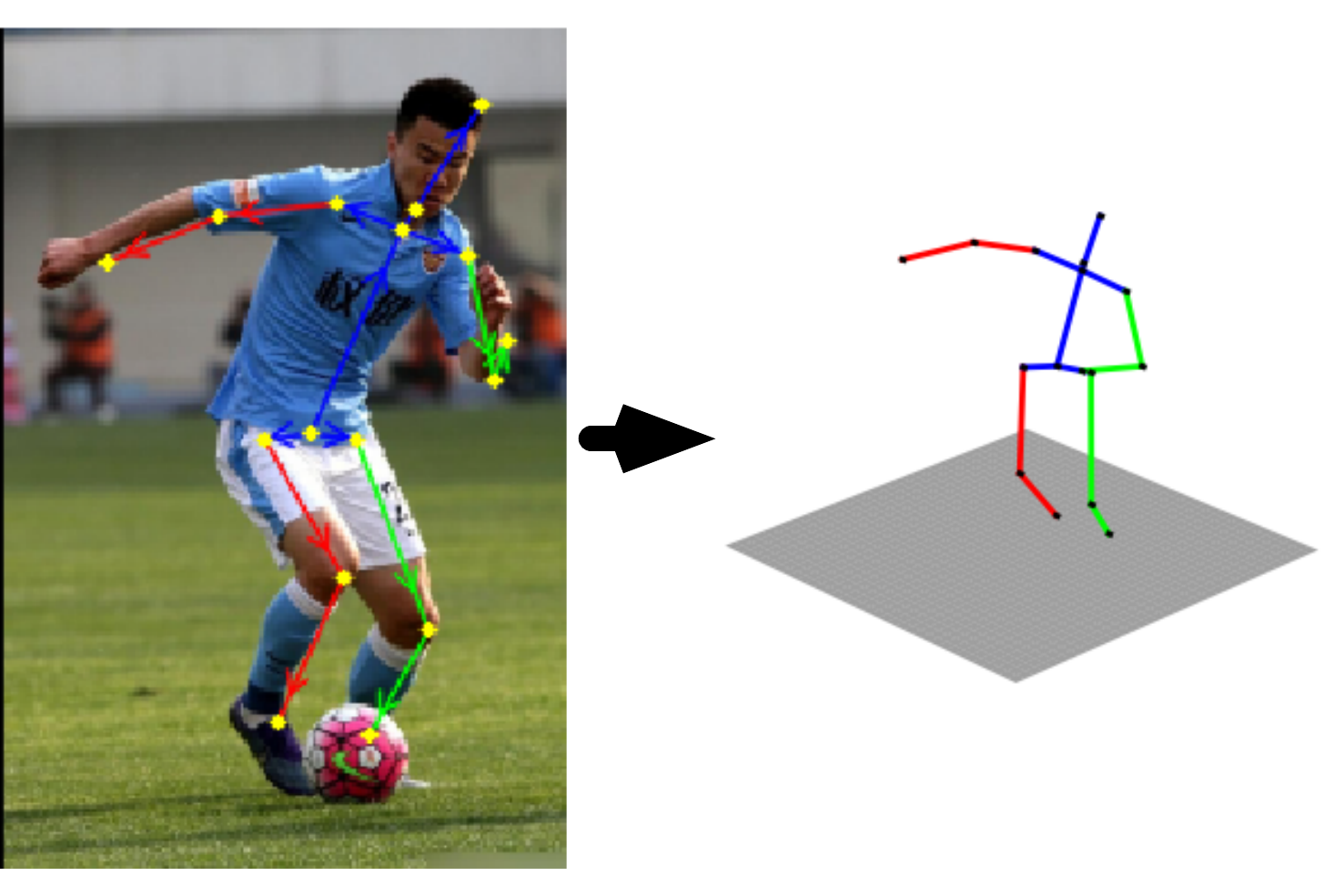}
  \caption{Input an image of a human, our approach firstly locates its 2D joints and predicts the Forward-or-Backward Information (FBI) for each bone. And then, the 3D pose is estimated by considering both of the two terms. }\label{fig:teaser}
\end{figure}

Reasoning 3D human pose from a single RGB image in the wild has drawn great attentions in the past three decades due to its broad application scenarios, such as autonomous driving, virtual reality, human-computer interaction and video surveillance. Recently, leveraging on large-scale well-annotated datasets such as MPII \cite{andriluka20142d}, significant progresses have been made in 2D human pose estimation using Convolutional Neural Networks(ConvNets)\cite{toshev2014deeppose,wei2016convolutional,newell2016stacked,chu2016structured, yang2016end, chu2017multi}. Although ConvNet models can be directly used for fitting the function $f$ that maps an image $I$ of a human to the 3D positions of the human skeleton joints $y$ as proposed in \cite{li20143d}, challenges remain for two reasons: 1) preparing such in-the-wild images dataset is extremely difficult whether be it by capturing using 3D sensors or annotating manually; 2) $f$ is hyper nonlinear and is hard to approximate.

In order to tackle this problem, state-of-the-art methods, such as \cite{moreno20173d,martinez2017simple, fang2018learning}, usually separate the mapping $y = f(I)$ into two functions and try to fit them using ConvNets individually. The two functions are the mapping from the image to 2d joint locations $x = g(I)$ and the mapping from 2D joint locations to 3D joint positions $y = h(x)$. They are connected by a compound operation and we have $y = h(g(I))$. This strategy greatly reduces the difficulty in obtaining annotated ground-truth data mainly in two aspects: Firstly, due to the ease of annotating 2D joints, it is possible to have a large amount of labeled images (e.g. MPII\cite{andriluka20142d}) for training a deep network to approximate the function $g$. Secondly, due to the flexibility of projecting a 3D pose using arbitrary viewpoints, it is also possible to have a large amount of synthetically generated 2D-to-3D pose pairs (as shown in \cite{fang2018learning}) for training a deep network to approximate the function $h$. Moreover, this two-step framework also alleviates the difficulty of fitting $f$ by decomposing it into two easier tasks of fitting $g$ and $h$ independently.

However, such a two-step framework is fundamentally flawed by oversimplifying the 3D pose estimation problem. Firstly, due to the weak supervision of 2D annotated poses, this approach causes loss of 3D-aware features during learning procedure. On the other hand, recovering 3D poses from 2D joint locations only is an ill-posed problem. Ambiguity exists since different yet valid 3D poses can explain the same observed 2D joints. In other words, although decomposing the learning procedure into two independent phases makes data annotation and function learning tractable, it produces a gap between the 2D human image and its corresponding 3D pose.

In this paper, we attempt to bridge this gap by involving Forward-or-Backward Information (FBI) of each bone associated with pair-wise connecting joints. Taking a forearm as example, the information whether this bone is forward or backward shows which end-joint on it (say, the elbow or the wrist) is closer to the view of camera. As validated in \cite{taylor2000reconstruction, zhou2010parametric}, the 3D joint positions can be uniquely determined by their 2D locations plus such FBI if taking the length ratios of bones as prior knowledge. In detail, our method decomposes the mapping $y = f(I)$ into three sub-functions: the mapping from the input image to 2D joint locations $g(I)$, the mapping from the input image to FBI $g'(I)$, and the mapping from these two terms to 3D coordinates of the joints $y = h(g(I), g'(I))$. In our work, all these sub-functions are approximated using deep neural networks. To support the training, we randomly selected 12K in-the-wild images from MPII\cite{andriluka20142d} and annotated the FBI for them using a well-designed user interface. Compared with previous works, our approach shows advantages in three manifolds: At first, using FBI as a kind of 3D-aware supervision to fit $g'$ digs out more effective information for predicting 3D human poses. Secondly, taking both 2D joint locations and FBI as input greatly reduces the ambiguity of 3D joints lifting. Thirdly, to label the FBI of an image, the annotators only need to do a binary selection for each bone. This makes it possible to build an infinitely large training dataset. Fig ~\ref{fig:teaser} shows an example of our 3D pose estimation.

In summary, our major contributions in this paper are:
\vspace{-3mm}
\begin{itemize}
\setlength\itemsep{-0em}
  \item the first work to exploit Forward-or-Backward Information (FBI) of bones for 3D human pose estimation, with which our method outperforms all previous works.
  \item we design a novel deep learning architecture that consists of two ConvNets to map an image of a human to the 2D joint locations and the FBI of all bones individually and a deep regression network to predict 3D joint positions using both 2D joint locations and FBI.
  \item we have labeled the FBI for 12K in-the-wild images with a well-designed user interface. They will be released to public to benefit other researchers working in this area.

\end{itemize}

\begin{figure*}[t]
  \centering
  \includegraphics[width=0.98\textwidth]{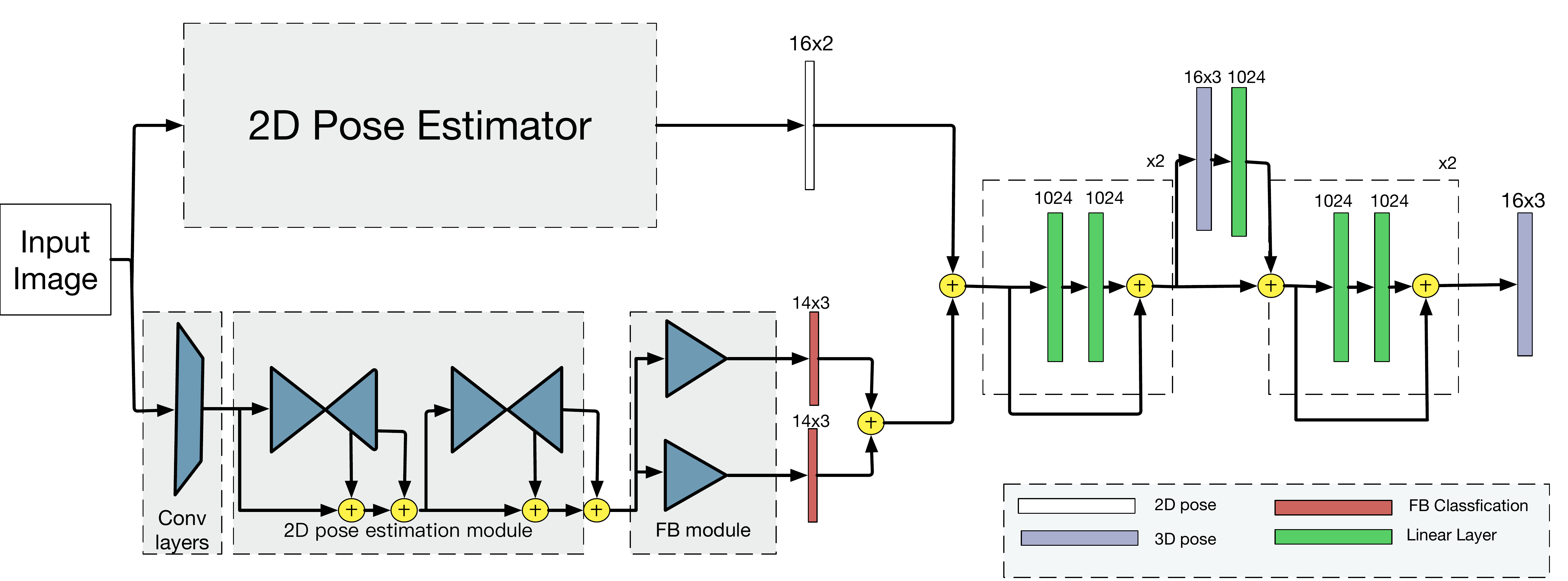}
  \caption{The network architecture of our method. It consists of a 2D pose estimator, a FBI classifier and a 3D pose regressor. $\bigoplus$ means concatenation.}\label{fig:network}
\end{figure*}

\section{Related works}
To capture human pose from a single image is a long-standing research topic in computer vision, readers can refer to \cite{zhang2016survey} and \cite{sarafianos20163d} for literature reviews of 2D and 3D human pose estimation respectively. In this paper, we only give a review of the algorithms using deep nets for 3D human pose prediction.
\vspace{-3mm}
\paragraph{Inferring 3D Pose by Encoder-Decoder} With a dataset of human images and their well-labeled 3D skeleton joints, Li and Chan\cite{li20143d} trained a deep ConvNets for image encoding and followed by two deep regression networks for 3D pose prediction and body part detection simultaneously. The algorithm in \cite{tekin2016structured} firstly trained an auto-encoder to represent the 3D joints into a high-dimensional space and then learned a ConvNets to map the input image to this space for 3D pose generation. To decode the 3D coordinates of the joints, instead of using a regression method, Pavlakos et al. \cite{pavlakos2017coarse} represented 3D joints in a volume and utilized a set of 3D deconvolutional layers for pose prediction in a coarse-to-fine way. Because of the challenges to obtain 3D groundtruth poses, these methods are difficult to be applied on in-the-wild images.
\vspace{-3mm}
\paragraph{Inferring 3D Pose by 2D Joints Estimation} To avoid collecting 2D-3D paired data in the wild, a large portion of recent works (such as \cite{moreno20173d, martinez2017simple, fang2018learning, zhou2016sparseness, chen20173d, bogo2016keep, mehta2017vnect, tome2017lifting, nie2017monocular}) decomposed the task of 3D pose inference into two independent stages: generating 2D poses firstly and then lifting them into 3D space. For example, Martinez et al. \cite{martinez2017simple} proposed to directly regress the 3D coordinates of the joints from their 2D locations with a sequence of fully connected layers. The method is very simple yet achieves state-of-the-art performance. Fang et al. \cite{fang2018learning} conducted 3D pose lifting from 2D joints by considering the prior knowledges of the relationships in-between skeleton joints. By exploiting geometric constraints, Zhou et al. \cite{zhou2017towards} proposed a weakly-supervised approach making these two stages possible to be trained together using in-the-wild images which have no 3D pose groundtruth. Most recently, adversarial learning framework was adopted in \cite{yang20183d} to ensure the anthropometrical validity of the output pose and further improved the performance. Our approach follows such two-stage framework but differs from it in two folds: 1) our first stage not only outputs the 2D joints but also the FBI; 2) correspondingly, the second stage takes 2D joints as well the FBI as input for 3D pose prediction.
\vspace{-3mm}
\paragraph{Building Training Dataset} There exist some works attempting to obtain 3D groundtruth poses for images to build training dataset. H3.6M \cite{ionescu2014human3} is one of such works in which the 3D pose of a human is captured using a Mocap system within an indoor environment. As a result, the learning-based methods trained with this dataset are difficult to be generalized to images in-the-wild. The work of \cite{mehta2017monocular}, therefore, captured actors in a green screen studio and synthesized new images by composting the segmented foregrounds with arbitrary backgrounds. Both \cite{chen2016synthesizing} and \cite{varol2017learning} took use of graphics methods to generate synthetic 3D human bodies and created images by overlaying them on a real background photo. Rogez et al.\cite{rogez2016mocap} proposed an image-based synthesis approach that firstly searched appropriate images according to a known 3D pose and then stitched them together. Although unlimited images can be synthesized using these methods, they still have very different appearances as real photos. This causes the challenges of domain adaptation to remain. Another kind of solutions for data collection is designing interactive annotation tools. The work called Poselets from Bourdev and Malik \cite{bourdev2009poselets} belongs to this category. However, the method shows extremely heavy user interventions and restricts the ability to construct a large dataset. In this work, we propose to annotate the FBI of bones instead of the 3D positions of joints. This change helps to reach a balance by not only reducing the amount of user interactions but also improving the performance by involving an extra supervision.

\section{Methodology}
Given an image of a human $I$, we represent the 2D pose as a set of skeleton joints $J_{2d} = \{p_1, p_2,...,p_n \}$ where $n$ denotes the number of joints ($n=16$ in this paper) and $p_i = (x_i, y_i)$ means the pixel the $i_{th}$ joint located at. Accordingly, the 3D pose is denoted as $J_{3d}=\{ P_1, P_2, ..., P_n\}$ and $P_i=(X_i, Y_i, Z_i)$ means the 3D coordinates (in this work, we use the coordinate system of the camera) of the $i_{th}$ joint located at. The 3D bones can be represented as $B_{3d}=\{ B_1, B_2, ..., B_m\}$ where $m$ is the number of bones(it equals $14$ in this work). $B_i$ indicates a directed vector which starts from one of its end joint $B^{0}_i$ to the other one $B^{1}_i$. The order of end joints associated with each bone is fixed and highlighted using an arrow as illustrated in Fig. ~\ref{fig:teaser}. So as to construct the mapping from $I$ to $J_{3d}$, our approach is built upon a ConvNets based deep learning architecture. Our methodology is introduced in the following three parts.

\subsection{FBI annotation}

\paragraph{What is FBI?} Each bone $B_i = \overrightarrow{B^{0}_iB^{1}_i}$, w.r.t the camera view, has three states: forward, backward and parallel to sight. This shows the depth order of $B^{0}_i$ and $B^{1}_i$ w.r.t camera frame reference. Therefore, the FBI of an image can be expressed using a binary matrix $F = \{f_1, f_2, ...,f_m\}$ where $f_i$ is a one-hot 3-dimensional vector, i.e., $f_{i}(j) = 1$ means the $i_{th}$ bone has the $j_{th}$ status and $j=0, 1, 2$.

\vspace{-3mm}
\paragraph{Annotating procedure} We recruited 6 paid annotators for FBI labeling. The images for annotation are randomly selected from MPII dataset \cite{andriluka20142d} where the 2D bones (projection of $B_{3d}$) are provided. Using our developed user interface, annotators were shown images one by one with the associated 2D bones overlayed on them (as illustrated in Fig. ~\ref{fig:teaser}). The bones are also shown one by one with highlight each time. For each one, the annotator is asked to make a choice from three options: forward, backward or uncertain. Considering the difficulties to give an accurate judgement whether a bone is "parallel to sight" or not, we replace this option by "uncertain". On average, one image can be annotated in 20 seconds. Note that, we randomly chose a set of images from H3.6M\cite{ionescu2014human3} and mixed them into the images for labeling. These images are shown to annotators in a random order. Based on 3D groundtruth poses of these images, the whole labeling procedure is monitored automatically: the annotator would be given a feedback when his or her labeled FBI is conflicting with the groundtruth. This greatly helps the annotators to learn how to make a good annotation and ensures the quality of labeled information.

\begin{figure}
	\begin{centering}
		\includegraphics[width=0.48\textwidth]{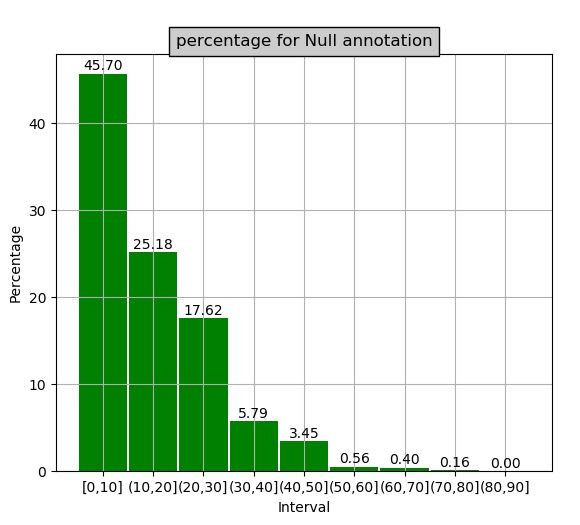}
	\end{centering}
	\caption{The distributions of out-of-plane angles for all bones marked as "uncertain". Each bar shows the percentage of the bones whose out-of-plane angle lies in a certain range.}
	\label{fig:distribution}
\end{figure}

\paragraph{Our FBI-Dataset} In all, we successfully annotated the FBI for around 12K in-the-wild images from MPI\cite{andriluka20142d}. Among them, around 20\% bones are marked as uncertain. We illustrate the distributions of out-of-plane angles for all uncertain bones in Fig ~\ref{fig:distribution}, from which we can see that people show more uncertainty when the bone is closer to parallelling with the view plane. Both our dataset and the annotating user interface will be released to public.

\subsection{Network architecture}
In short, our network consists of three components: a 2D pose estimator, a FBI predictor and a 3D pose regressor. Our 2D pose estimator used the same architecture as in \cite{andriluka20142d} aiming to take an image $I$ of a human as input and output the 2D locations of all 16 joints of the human in $I$. The detailed design of other two components are described as follows.

\paragraph{FBI Predictor} This module inputs an image $I$ and outputs the FBI of all 14 bones, where $f_i$ can have three statuses: forward, backward or uncertain. This problem thus is formulated as a per-bone classification task. Taking $I$ as input, our network for this component starts from a sequence of convolutional layers that are followed by two successive stacked hourglass modules(refer to \cite{andriluka20142d} for the design of such modules). The extracted feature maps are then fed into a set of convolutional layers. Finally, these are followed by a fully connected layer with a softmax layer to output classification results. The number of neurons for these layers are set as same as in \cite{zhou2017towards}.

\paragraph{3D Pose Regressor} In our work, we learn a deep regression network to infer the 3D coordinates of the joints by taking both their 2D locations and the FBI as input. Instead of using the discrete $F$, this regressor takes the generated probability matrix of the softmax layer as input which provides more information. We first concatenate the 2D locations and the probability matrix together and then map them to the 3D pose by exploiting two cascaded blocks as used in \cite{fang2018learning}. Specifically, each block maps the input feature into a higher dimension using two fully connected layers (1024 neurons are used in our work) interleaved with Batch Normalization, Dropout layers, and ReLU activation. At the end of the first block, an extra linear layer is utilized to output a coarse 3D pose. This is further re-projected into a 1024-dimension space and serves as a part of the input for the second block. We apply residual connections between the two blocks for sake of taking fully use of the information.

\paragraph{Weighted FBI Supervision} By observing the FBI labeling procedure, we found that it shows high difficulty to separate the forward(or backward) status from uncertain one. In other words, these two statuses have no sharp gap. This heavily limits us to extract proper features due to the blurry supervision. To deal with this problem, we propose to use two kinds of weighted FBI supervision to extract features with different focuses and then combine them together for 3D pose regression. We first use a fixed weighting strategy and assign different weights for the bones with different states when doing FBI classification. In this work, $1$ is set for the bones with forward and backward status and $0.05$ is set for the bones with uncertain status. Such weighting generates a probability map $P_{fws}$ (fws is a shorthand for fixed weighting supervision) that mostly captures the information of the bones with large out-of-plane angle. Another weighting strategy we used is in an adaptive way that assigns dynamic weights for bones during the training procedure. This is implemented using focal loss \cite{lin2017focal}. For each bone $B_i$, we use a one-hot 3-dimensional vector $g_i$ to denote its groundtruth label and use a 3-dimensional vector $p_i$ to represent the output probability vector. Thus, the loss function for the FBI classification of $B_i$ has the following formula, where $\gamma$ is set to be 2 in our experiments.
\begin{equation}\label{eq:focaloss}
  Loss(p_i) = \sum^{2}_{j=0}(-(1 - p_i(j))^{\gamma}log(p_i(j)))*g_i(j)
\end{equation}

Intuitively speaking, using this method, a bone would have lower weight if it is already of a high probability to be classified into one category. We denote the output probability map using this strategy as $P_{aws}$ (aws is a shorthand for adaptive weighting supervision) which takes more care about the bones owning small out-of-plane angle. Our final 3D pose regressor takes both $P_{fws}$ and ${P_{aws}}$ as input. The efficiency of this strategy is validated in Section ~\ref{sec:ablation}. The whole network architecture of our method is illustrated in Fig ~\ref{fig:network}.

\subsection{Training}
Our training procedure includes three steps from local to global and from coarse to fine. At first, we train FBI predictor and 3D pose regressor respectively from scratch and use the pre-trained model provided by \cite{newell2016stacked} for the 2D pose estimator. In this step, the FBI predictor is trained using the images in H3.6M \cite{ionescu2014human3} where the 3D groundtruth poses are converted into FBI formula. Afterwards, the images from H3.6M are fed into both the trained 2D pose estimator and FBI predictor. We take their outputs and the 3D groundtruth poses as paired data to train the 3D pose regressor. As the second step, to reach a global optimal, all these three components are connected and the parameters are finetuned simultaneously with the training images in H3.6M. Finally, we finetune both the FBI predictor and the 3D pose regressor using our annotated FBI-dataset for generalizing our network into in-the-wild domain. To support this training, our approach is performed in a weakly-supervised way: the output of the 3D pose regressor is followed by a set of fully connected layers and a softmax layer for FBI classification where such layers are pre-trained using synthetic data.

\paragraph{Converting 3D pose into FBI formula} To support our training, a critical problem is how to convert the 3D groundtruth pose in H3.6M into a FBI formula. We exploit a thresholding method: for a bone, it is marked as forward if its out-of-plane angle is larger than $alpha$, backward if the angle is smaller than $-\alpha$ and uncertain for other cases. Intuitively, learning the FBI predictor would be getting harder if $\alpha$ is getting smaller while the useful information from uncertain bones would be getting fewer when $alpha$ is getting larger. In our work, $\alpha$ is set as $35^{\circ}$ to reach a balance based on our experiments in Section ~\ref{sec:ablation}.
\begin{table*}[]
\centering
\label{tab:pro1}
\begin{tabular}{lcccccccc}
\textbf{Protocol \#1}          & Direct        & Discuss       & Eating        & Greet         & Phone         & Photo         & Pose          & Purch.        \\ \hline
LinKDE\cite{ionescu2014human3}                & 132.7         & 183.6         & 132.3         & 164.4         & 162.1         & 205.9         & 150.6         & 171.3         \\
Tekin et al. \cite{tekin2016structured}            & 102.4         & 147.2         & 88.8          & 125.3         & 118.0         & 182.7         & 112.4         & 129.2         \\
Du et al. \cite{du2016marker}              & 85.1          & 112.7         & 104.9         & 122.1         & 139.1         & 135.9         & 105.9         & 166.2         \\
Chen \& Ramanan \cite{chen20173d}        & 89.9          & 97.6          & 89.9          & 107.9         & 107.3         & 139.2         & 93.6          & 136.0         \\
Pavlakos et al. \cite{pavlakos2017coarse}  & 67.4          & 71.9          & 66.7          & 69.1          & 72.0          & 77.0          & 65.0          & 68.3          \\
Zhou et al. \cite{zhou2017towards}     & 54.8          & 60.7          & 58.2          & 71.4        & 62.0 & \textbf{65.5} & 53.8          & 55.6          \\
Martinez et al. \cite{martinez2017simple} & 51.8          & 56.2          & 58.1          & 59.0          & 69.5          & 78.4          & 55.2          & 58.1          \\
Fang et al. \cite{fang2018learning}      & 50.1          & 54.3          & 57.0          & 57.1          & 66.6          & 73.3          & 53.4          & 55.7          \\
Ordinal \cite{pavlakos2018ordinal} & \textbf{48.5} & 54.4 & 54.4 & 52.0 &\textbf{59.4} & \textbf{65.3} & \textbf{49.0}  & 52.9 \\
\hline
Ours-Baseline         & 49.7          & 53.9          & 53.4          & 56          & 62.6          & 70.5          & 53.4          & 52.2          \\
Ours-WS             & 50.2          & 53.2          & 54.0          & 56.4          & 62.7          & 71.3          & 53.4          & 52.3          \\
Ours-Final    & \textcolor{blue}{\textbf{49.1}} &\textcolor{blue}{ \textbf{52.7}} & \textcolor{blue}{\textbf{52.0}} & \textcolor{blue}{\textbf{55.2}} & \textcolor{blue}{\textbf{60.5}}     & 69.8          & \textcolor{blue}{\textbf{52.3}} & \textcolor{blue}{\textbf{51.5}} \\ \hline
                               & Sitting       & SittingD.     & Smoke         & Wait          & WalkD.        & Walk          & WalkT.        & Average       \\ \hline
LinKDE\cite{ionescu2014human3}                 & 151.6         & 243.0         & 162.1         & 170.7         & 177.1         & 96.6          & 127.9         & 162.1         \\
Tekin et al. \cite{tekin2016structured}            & 138.9         & 224.9         & 118.4         & 138.8         & 126.3         & 55.1          & 65.8          & 125.0         \\
Du et al. \cite{du2016marker}                & 117.5         & 226.9         & 120.0         & 117.7         & 137.4         & 99.3          & 106.5         & 126.5         \\
Chen \& Ramanan \cite{chen20173d}         & 133.1         & 240.1         & 106.6         & 106.2         & 87.0          & 114.0         & 90.5          & 114.1         \\
Pavlakos et al. \cite{pavlakos2017coarse}         & 83.7          & 96.5          & 71.7          & 65.8          & 74.9          & 59.1          & 63.2          & 71.9          \\
Zhou et al. \cite{zhou2017towards}            & 75.2          & 111.6         & 64.1          & 66.0          & \textbf{51.4} & 63.2          & 55.3          & 64.9          \\
Martinez et al. \cite{martinez2017simple} & 74.0          & 94.6          & 62.3          & 59.1          & 65.1          & 49.5          & 52.4          & 62.9          \\
Ordinal \cite{pavlakos2018ordinal}& 65.8 & \textbf{71.1} & 56.6 & \textbf{52.9} & 60.9    & 44.7 & \textbf{47.8} & \textbf{56.2 }\\
Fang et al. \cite{fang2018learning}    & 72.8          & 88.6          & 60.3          & 57.7          & 62.7          & 47.5          & 50.6          & 60.4          \\ \hline
Ours-Baseline                           & 66.5          & 80.7          & 57.5         & 56          & 60.9          & 45.9          & 50.7          & 58.0          \\
Ours-WS                 & 64.7          & 77.0          & 57.4          & 55.5          & 61.0          & 45.6          & 50.4          & 57.7          \\
Ours-Final                 & \textcolor{blue}{\textbf{64.3}} & \textcolor{blue}{\textbf{74.8}} & \textcolor{blue}{\textbf{56.4}} & \textcolor{blue}{\textbf{55.1}} & 60.0          & \textcolor{blue}{\textbf{44.4}} & \textcolor{blue}{\textbf{48.9}} & \textcolor{blue}{\textbf{56.5}}
\end{tabular}
\caption{Quantitative comparisons of the mean per joint position error(MPJPE) between our prediction and the ground truth on Human3.6M under Protocol \#1. Note that, Ordinal \cite{pavlakos2018ordinal} is a concurrent work with our method. The best score without consideration of this work is marked in blue bold. We also use black bold to highlight the best score when taking this work for comparison.}
\end{table*}

\section{Experimental Results}
Our approach was implemented based on the code released by \cite{martinez2017simple} that uses tensorflow. During the first training stage described in Section 3.3, it took 100K iterations with a batch size of 8 to train the FBI predictor for both fixed weighting and adaptive weighting strategy based on the pretrained model from \cite{zhou2017towards}. The batch size of the second global finetuning stage is set as 8 and the procedure took 100K iterations. Our final FBI-predictor finetuning stage was conducted in 120K iterations by taking a batch size of 8. The whole training procedure took about two days in one Titan X GPU with CUDA 8.0 and cudnn 5. During testing phase, one forward passing takes 30ms. To verify the efficiency of our proposed algorithm, we conduct the evaluation in the following manifolds.

\subsection{Comparisons against existing methods}
Our numerical evaluation is conducted on Human3.6M \cite{ionescu2014human3}. As far as we know, this is one of the largest public dataset where both the 2D pose and 3D groundtruth pose are available. It contains 3.6 millions of RGB images captured by a MoCap System in an indoor environment, in which 7 professional actors performed 15 activities such as walking, eating, sitting, making a phone call and engaging in a discussion. The videos are down-sampled from 50fps to 10fps in order to reduce redundancy. Following the standard protocol as in [10, 17], we use 5 subjects(S1, S5, S6, S7, S8) for training and the rest 2 subjects(S9, S11) for evaluation. The mean per joint position error(MPJPE) between the ground truth and our prediction (after alignment of the central hip joint) is used as our evaluation metric. This is denoted as protocol \#1. For some of previous works, the prediction has been further aligned with the ground truth via a rigid transformation. This post-processing is named protocol \#2.

\begin{table*}[]
\centering
\label{tab:pro2}
\begin{tabular}{lcccccccc}
\textbf{Protocol \#2}          & Direct        & Discuss       & Eating        & Greet         & Phone         & Photo         & Pose          & Purch.        \\ \hline
Bogo et al. \cite{bogo2016keep}             & 62.0          & 60.2          & 67.8          & 76.5          & 92.1          & 77.0          & 73.0          & 75.3          \\
Moreno-Noguer \cite{moreno20173d}          & 66.1          & 61.7          & 84.5          & 73.7          & 65.2          & 67.2          & 60.9          & 67.3          \\
Pavlakos et al. \cite{pavlakos2017coarse}          & 47.5          & 50.5          & 48.3          & 49.3          & 50.7          & 55.2          & 46.1          & 48.0          \\
Martinez et al. \cite{martinez2017simple}        & 39.5          & 43.2          & 46.4          & 47.0          & 51.0          & 56.0          & 41.4          & 40.6          \\
Fang et al. \cite{fang2018learning}           & \textcolor{blue}{\textbf{38.2}} & 41.7   & 43.7          & 44.9      & 48.5          & 55.3          & 40.2          & 38.2 \\
Ordinal \cite{pavlakos2018ordinal}& \textbf{34.7} & \textbf{39.8} & \textbf{41.8} & \textbf{38.6} &\textbf{ 42.5} & \textbf{47.5} & \textbf{38.0} & \textbf{36.6} \\
\hline
Ours-Baseline           & 38.6          & 41.8          & 42.1          & 44.9         & 45.8      & 52.5          & 40.4          & 38.4          \\
Ours-WS                 & 38.9          & 41.5          & 43.0          & 45.2          & 46.2           & 52.4          & 40.6          & 38.3          \\
Ours-Final   & 38.4          & \textcolor{blue}{\textbf{41.3}}      & \textcolor{blue}{\textbf{41.9}}     & \textcolor{blue}{\textbf{44.7}}   & \textcolor{blue}{\textbf{45.4 }} & \textcolor{blue}{\textbf{51.7}} & \textcolor{blue}{\textbf{40.0}} & \textcolor{blue}{\textbf{37.9}} \\ \hline
                               & Sitting       & SittingD.     & Smoke         & Wait          & WalkD.        & Walk          & WalkT.        & Average       \\ \hline
Bogo et al. \cite{bogo2016keep}             & 100.3         & 137.3         & 83.4          & 77.3          & 86.8          & 79.7          & 87.7          & 82.3          \\
Moreno-Noguer \cite{moreno20173d}          & 103.5         & 74.6          & 92.6          & 69.6          & 71.5          & 78.0          & 73.2          & 74.0          \\
Pavlakos et al. \cite{pavlakos2017coarse}       & 61.1          & 78.1          & 51.1          & 48.3          & 52.9          & 41.5          & 46.4          & 51.9          \\
Martinez et al. \cite{martinez2017simple}       & 56.5          & 69.4          & 49.2          & 45.0          & 49.5          & 38.0          & 43.1          & 47.7          \\
Fang et al. \cite{fang2018learning}            & 54.5          & 64.4          & 47.2          & 44.3          & 47.3          & 36.7          & 41.7          & 45.7          \\
Ordinal \cite{pavlakos2018ordinal}& 50.7 & \textbf{56.8} & \textbf{42.6} & \textbf{39.6} & \textbf{43.9} & \textbf{32.1} &\textbf{36.5} & \textbf{41.8} \\ \hline
Ours-Baseline                           & 50.2          & 63.1          & 45.6 & 42.2          & 46.3 & 34.1          & 39.3          & 44.3          \\
Ours-WS                 & 49.3 & 60.2          & 45.6          & 41.9          & 46.7          & 34.1          & 39.5          & 44.2          \\
Ours-Final    & \textcolor{blue}{\textbf{49.1}}          & \textcolor{blue}{\textbf{59.5}} & \textcolor{blue}{\textbf{45.2}}         & \textcolor{blue}{\textbf{41.9}} & \textcolor{blue}{\textbf{46.2} }  & \textcolor{blue}{\textbf{33.6}} & \textcolor{blue}{\textbf{38.8}} & \textcolor{blue}{\textbf{43.7}}
\end{tabular}
\caption{Quantitative comparisons of the mean per joint position error(MPJPE) between our prediction and the ground truth on Human3.6M under Protocol \#2. Note that, Ordinal \cite{pavlakos2018ordinal} is a concurrent work with our method. The best score without consideration of this work is marked in blue bold. We also use black bold to highlight the best score when taking this work for comparison. }
\end{table*}

The comparison results on protocol \#1 are reported in Table ~\ref{tab:pro1} while the results on protocol \#2 are shown in Table ~\ref{tab:pro2}. As seen from the table, our method outperforms all previous works almost on all actions. It is worth mentioning that our approach makes considerable improvements on some complicated actions like “sitting” and “sitting down” which show more challenges than others. Thanks to the FBI information, our method is very good at for the cases with large poses. Before our work, the best performance is achieved by Fang et al. \cite{fang2018learning} that involved the relationship priors between skeleton joints into the learning procedure. Compared with this, our results are 4mm more accurate on average. And we also believe their strategy can be integrated into our framework to get further improvement. It is worth mentioning that, concurrently, the work in \cite{pavlakos2018ordinal} exploited a similar strategy as ours and achieved comparable results which are also shown in Table ~\ref{tab:pro1} and Table ~\ref{tab:pro2}. Specifically, it proposed an annotation tool for collecting the depth relations for all joints. Such ordinal depth information is further used for training a neural network in a weakly supervised way. Comparing with this work, our annotation procedure is much easier for two reasons: 1) using our tool, annotators only need to mark the ordinal depth information for two joints with rigid connection (as a bone). This is more intuitive than the task of marking the depth relations for two separate joints as in \cite{pavlakos2018ordinal}. 2) for each image, to obtain the global depth orders of all joints, the annotation in \cite{pavlakos2018ordinal} usually requires the annotators to answer dozens of questions while ours has only 16 questions to be answered. In addition to, Yang et al.\cite{yang20183d} proposed another concurrent method which applied an adversarial learning framework and reached comparable performance as ours. It is convenient to combine their idea into our framework and we believe this will produce better results.

\begin{figure*}
  \centering
  \includegraphics[width=0.98\textwidth]{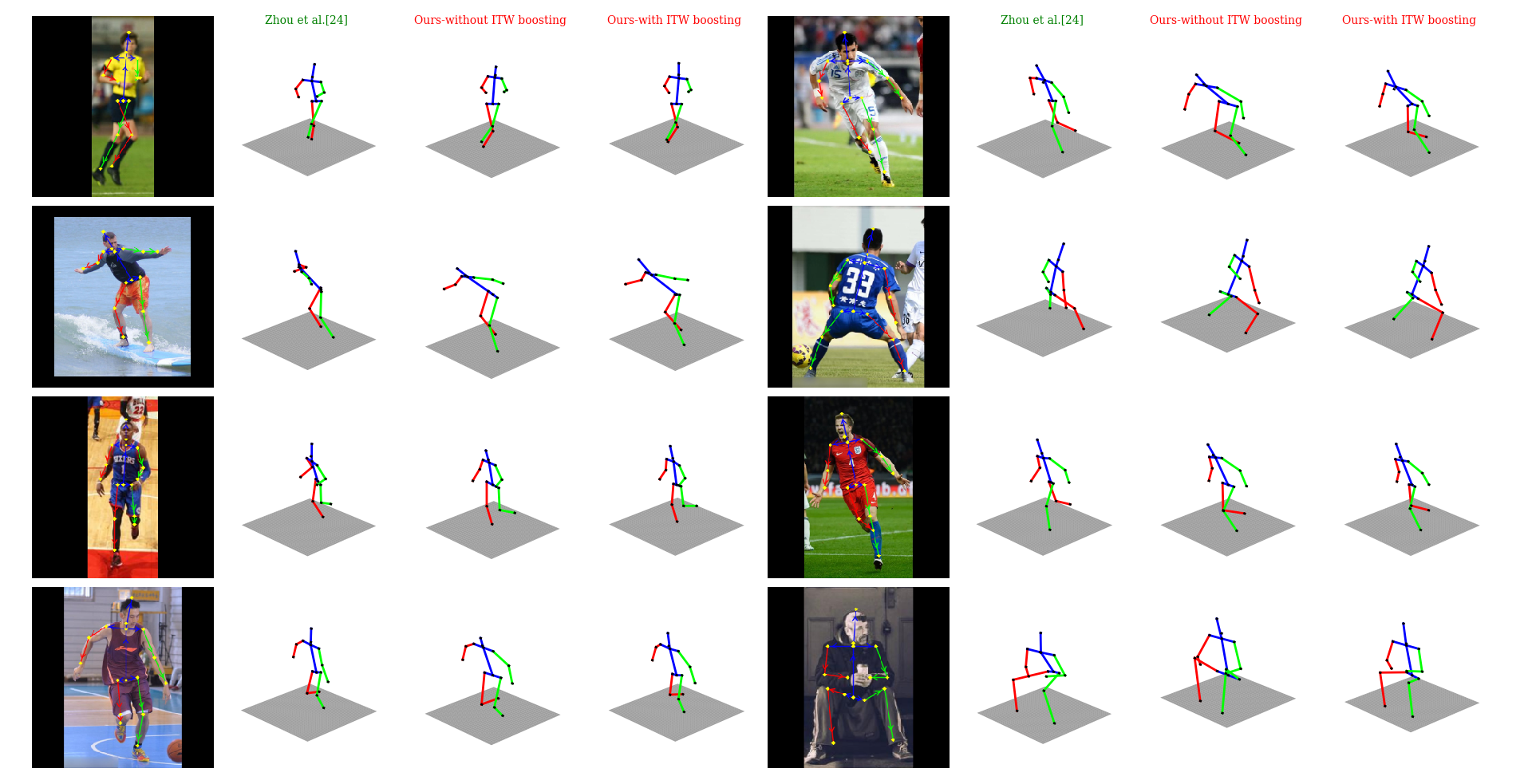}
  \caption{Quantitative comparison results of our method against others on some in-the-wild images which are chosen from MPII\cite{andriluka20142d}. ITW is shorthand for In The Wild.}\label{fig:wild}
\end{figure*}

\paragraph{Comparisons on in-the-wild generalization}
To validate the efficiency of our in-the-wild generalization, we also conduct comparisons against previous works on images in the wild both qualitatively and quantitatively. To our best knowledge, there is no dataset owning 3D pose groundtruth. To support the evaluation, we take 1K images from our FBI-dataset as the test data. We take the method \cite{zhou2017towards} to attend this comparison because of its state-of-the-art performance and accessible code. With each method, the 3D pose results are generated for all the 1K images firstly and then we use the correctness ratio of FBI derived from the 3D pose as the evaluation metric. Here, we only do the statistic on the bones with clear status, say forward or backward. The method of \cite{zhou2017towards} has 75\% correctness ratio while ours reaches 78\%. This is also verified by some qualitative comparison results as shown in Fig ~\ref{fig:wild}.

\subsection{Ablation studies}
\label{sec:ablation}
We study the influence on final performance of different choices made in our network design and the training procedure. The method without weighted supervision and in-the-wild FBI finetuning is denoted as "Ours-baseline" representing a baseline of our approach. Specifically, it used equal weighted loss for the training of FBI classification. Adopting weighted FBI supervision on this baseline method gives rise to "Ours-WS". 
After that, the strategy to exploit our in-the-wild FBI dataset for network finetuning is applied on "Ours-WS" and derives our final method "Ours-Final".

\begin{figure}
	\begin{centering}
		\includegraphics[width=0.48\textwidth]{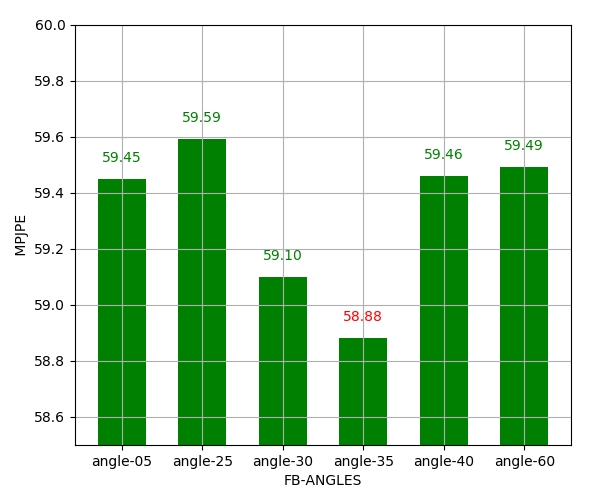}
	\end{centering}
	\caption{Our performance changes with the thresholding angle selection. All the values are obtained by using "Ours-baseline" on H3.6M protocol \#1.}
	\label{fig:bestangle}
\end{figure}

\paragraph{With/without weighted supervision} We first study the effectiveness of our weighted FBI supervision strategy by evaluate "Ours-Baseline" and "Ours-WS". All these methods are evaluated using the same way as in Section 4.1 and the results are also presented in Table ~\ref{tab:pro1} and Table ~\ref{tab:pro2}. It is not difficult to find that "Ours-WS" shows better performances than "Ours-baseline" for the actions with large poses (e.g. sitting down) and also keeps comparable accuracy for other actions. This validated the effectiveness of the design with two kinds of weighting supervision.

\paragraph{With/without in-the-wild FBI boosting}
Without our FBI-dataset, our approach ("Ours-WS") also outperforms all existing methods. However, the model is only trained on images from H3.6M and would have low performance on in-the-wild images. To overcome this problem, we use a large amount of images with labeled FBI to finetune our model. As shown in Table ~\ref{tab:pro1} and Table ~\ref{tab:pro2}, with such finetuning, our final method also has a performance increasing based on the evaluation on H3.6M. We also perform the comparison between the method with and without finetuning on in-the-wild generalization as did in Section 4.1, where our method without in-the-wild boosting only produces 58\% correct rate. This is also validated by some qualitative results as illustrated in Fig ~\ref{fig:wild}.

\paragraph{The best thresholding to convert 3D pose into FBI} As mentioned in Section 3.3, a critical issue for converting the 3D pose into FBI representation is determining the threshold angle $alpha$. To make a better choice, we generate several samples as ($5^{\circ},20^{\circ}, 25^{\circ}, 30^{\circ}, 35^{\circ}, 40^{\circ}, 60^{\circ}$) and conduct evaluation on all of them individually. The curve representing changes of accuracy along with $alpha$ is shown in Fig ~\ref{fig:bestangle}, from which we found that $35^{\circ}$ leads to the best performance.

\section{Conclusion}
In this paper, we exploited a new information called Forward-Backward Information(FBI) of bones targeting an end-to-end framework to generate 3D human poses from a single RGB image. The biggest challenge in this area is the lack of images dataset with 3D groundtruth poses. The previous works solve this problem by decomposing the task into two stages: performing 2D pose estimation and inferring 3D pose from only the 2D joints. These two stages are usually treated separately while incurs a gap between images and 3D poses. The involving of FBI tackles this issue in three ways: 1) with FBI supervision, we dig out more 3D-aware features from images; 2) taking both FBI and 2D joints as input to infer the 3D pose greatly reduces the ambiguity; 3) more importantly, the FBI is very easy to annotate where we marked such information for around 12K images which are keeping updated. The experiments demonstrate the effectiveness of our approach both qualitatively and quantitatively. We also believe our proposed approach can inspire others to involve some weakly yet easily annotated information for weakly supervised learning.

\section*{Acknowledgement}
The work was partially supported in part by Shenzhen Fundamental Research Fund under Grant No. KQTD2015033114415450, National Natural Science Foundation of China (grant no.: 61771201) and the Program for Guangdong Introducing Innovative and Enterpreneurial Teams (Grant No.: 2017ZT07X183).

{\small
\bibliographystyle{ieee}
\bibliography{egbib}
}

\end{document}